\documentclass[conference]{IEEEtran}
\IEEEoverridecommandlockouts

\usepackage{cite}
\usepackage{amsmath,amssymb,amsfonts,bm}
\usepackage[hyphens]{url}
\usepackage{algorithmic}
\usepackage{graphicx}
\usepackage{textcomp}
\usepackage[dvipsnames,svgnames,x11names]{xcolor}

\definecolor{repairblue}{HTML}{2563EB}
\definecolor{corruptred}{HTML}{DC2626}
\definecolor{cleangreen}{HTML}{16A34A}
\definecolor{scoreorange}{HTML}{EA580C}
\definecolor{gridbg}{HTML}{F1F5F9}

\usepackage{tikz}
\usetikzlibrary{
  positioning, arrows.meta, calc,
  decorations.pathreplacing, shapes.geometric,
  backgrounds, fit
}
\usepackage{pgfplots}
\usepgfplotslibrary{groupplots}
\pgfplotsset{compat=1.18}

\usepackage{booktabs}
\usepackage{array,tabularx}
\usepackage{multirow}
\usepackage{colortbl}

\newcommand{\uniformtable}{%
  \footnotesize
  \renewcommand{\arraystretch}{1.08}%
  \setlength{\tabcolsep}{4pt}%
}

\usepackage[capitalize,nameinlink]{cleveref}

\newcommand{\jure}{\textsc{JuRe}}
\newcommand{\RR}{\mathbb{R}}
\newcommand{\EE}{\mathbb{E}}

\AtBeginDocument{}

\begin{document}

% OLD: \title{Back to Repair: A Minimal Denoising Network for\\Time Series Anomaly Detection}
\title{Just Repair: A Minimal Denoising Network for\\Time Series Anomaly Detection}

\author{%
  Kadir\mbox{-}Kaan \"Ozer\IEEEauthorrefmark{1,}\IEEEauthorrefmark{2},
  Ren\'{e} Ebeling\IEEEauthorrefmark{1},
  Markus Enzweiler\IEEEauthorrefmark{2}%
  \thanks{%
    \IEEEauthorrefmark{1}Mercedes-Benz AG, Germany,
    \{kadir.oezer, rene.ebeling\}@mercedes-benz.com\par
    \IEEEauthorrefmark{2}Institute for Intelligent Systems,
    Esslingen University of Applied Sciences, Germany,
    \{markus.enzweiler\}@hs-esslingen.de%
  }%
}

\maketitle

\begin{abstract}
Time series anomaly detectors have grown steadily more complex, incorporating attention mechanisms, adversarial training, and stochastic latent variables. Yet, it is unclear how much of this machinery detection actually requires. We test this question with \jure{} (Just Repair), a deliberately minimal detector: a single depthwise-separable convolutional residual block trained to repair Gaussian-corrupted, channel-masked windows, scored at inference by a fixed structural discrepancy function with no learned parameters.
% OLD: \jure{} ranks second on the TSB-AD multivariate benchmark (AUC-PR 0.404 over 180 series) and second on the UCR univariate archive (AUC-PR 0.201 over 250 series), where it leads all neural baselines, while running roughly $20\times$ faster than the top-ranked method.
\jure{} ranks second on the TSB-AD multivariate benchmark (AUC-PR 0.404 over 180 series) and second on the UCR univariate archive (AUC-PR 0.201 over 250 series), where it leads all neural baselines. On TSB-AD, \jure{} runs roughly $20\times$ faster than AxonAD, one of the top-ranked methods on that benchmark.
% OLD: Full-benchmark ablations locate the source of this performance in the training objective rather than the architecture: removing Gaussian corruption costs 0.045 AUC-PR, whereas architecture variants at matched parameter budgets differ by at most 0.017.
Full-benchmark ablations show that removing Gaussian corruption reduces AUC-PR by 0.046, whereas AUC-PR across the evaluated architecture variants spans at most 0.017.
% OLD: A synthetic manifold experiment with known ground truth supports the underlying geometric interpretation, showing anomaly scores that correlate with true off-manifold distance (Pearson $r=0.725$) and repair directions that align increasingly with the true projection as anomaly magnitude grows.
A synthetic linear-manifold experiment provides partial evidence for this geometric interpretation: anomaly scores correlate with true off-manifold distance (Pearson $r=0.725$), and repair directions align increasingly with the true projection as anomaly magnitude grows.
% OLD: Wilcoxon signed-rank tests with Holm correction establish significance against 20 of 25 baselines. Code is available at \url{https://github.com/iis-esslingen/JuRe}.
Wilcoxon signed-rank tests with Holm correction find significant differences against 20 of 25 baselines, although dependence among series limits dataset-level interpretation. Code is available at \url{https://github.com/iis-esslingen/JuRe}.
\end{abstract}


\section{Introduction}\label{sec:intro}

% OLD: Under the manifold hypothesis, time series anomaly detection is a geometric problem: normal behaviour occupies a low-dimensional manifold in window space, and anomalies fall off it. The Bayes-optimal denoiser under Gaussian corruption recovers the posterior mean, which for small noise levels approximates projection onto this manifold~\cite{vincent2011connection}. Training a network on corrupted inputs therefore teaches it the geometry of normal data without explicit density estimation.
Under the manifold hypothesis, normal time-series windows concentrate near a lower-dimensional set in window space, and anomalies depart from that set. The Bayes-optimal denoiser under Gaussian corruption recovers the posterior mean; at small noise levels, its correction is related to the score of the smoothed data density~\cite{vincent2011connection}. This result motivates learning denoising corrections from normal data without explicit density estimation, but it does not guarantee projection by a finite neural network on arbitrary anomalous inputs.
% OLD: Denoising autoencoders exploit this property implicitly~\cite{vincent2008extracting,vincent2010stacked}, the score-matching literature formalises it~\cite{song2020generativemodelingestimatinggradients}, and recent work shows that predicting clean, low-dimensional targets enables far smaller architectures than previously thought~\cite{li_jit_2025}.
Denoising autoencoders exploit this property implicitly~\cite{vincent2008extracting,vincent2010stacked}, the score-matching literature formalizes it~\cite{song2020generativemodelingestimatinggradients}, and recent work shows that predicting clean, low-dimensional targets enables far smaller architectures than previously thought~\cite{li_jit_2025}.

% OLD: The anomaly detection literature has moved in the opposite direction. Association discrepancy over attention~\cite{xu_anomaly_2022}, predictive query dynamics~\cite{ozer_axonad_2026}, dual-path variational encoders~\cite{ozer_stream-vae_2025}, stochastic recurrent networks~\cite{su_robust_2019}, and adversarial meta-learning~\cite{tuli_tranad_2022} each add substantial learned machinery---sidestepping iterative diffusion~\cite{tashiro2021csdiconditionalscorebaseddiffusion} only to introduce complexity of a different kind. Whether this complexity buys genuine detection quality over a principled repair baseline has, to our knowledge, never been tested systematically on standardized benchmarks.
The anomaly detection literature has moved in the opposite direction. Association discrepancy over attention~\cite{xu_anomaly_2022}, predictive query dynamics~\cite{ozer_axonad_2026}, dual-path variational encoders~\cite{ozer_stream-vae_2025}, stochastic recurrent networks~\cite{su_robust_2019}, and adversarial meta-learning~\cite{tuli_tranad_2022} each add substantial learned machinery. They sidestep iterative diffusion~\cite{tashiro2021csdiconditionalscorebaseddiffusion} only to introduce complexity of a different kind. Whether this complexity buys genuine detection quality over a principled repair baseline has, to our knowledge, never been tested systematically on standardized benchmarks.

\begin{figure}[t]
\centering
\resizebox{0.75\columnwidth}{!}{%
\begin{tikzpicture}[
    x={(-0.866cm,-0.5cm)},
    y={(0.866cm,-0.5cm)},
    z={(0cm,1cm)},
    scale=1.35,
    every node/.style={font=\small}
]
  \draw[-{Stealth[length=5pt]}, gray!50, thick] (0,0,0) -- (5.3,0,0);
  \draw[-{Stealth[length=5pt]}, gray!50, thick] (0,0,0) -- (0,5.3,0);
  \draw[-{Stealth[length=5pt]}, gray!50, thick] (0,0,0) -- (0,0,3.0)
    node[anchor=west, text=gray!80, font=\small, yshift=-4pt] {$\mathbb{R}^{W \times C}$};

  \filldraw[cleangreen!10, draw=cleangreen, thick, fill opacity=0.85, line join=round]
    plot[domain=1:5, samples=20] (\x, 1, {1 + 0.1*(\x-3)^2 - 0.1*(1-3)^2 + 0.2*1}) --
    plot[domain=1:5, samples=20] (5, \x, {1 + 0.1*(5-3)^2 - 0.1*(\x-3)^2 + 0.2*\x}) --
    plot[domain=5:1, samples=20] (\x, 5, {1 + 0.1*(\x-3)^2 - 0.1*(5-3)^2 + 0.2*5}) --
    plot[domain=5:1, samples=20] (1, \x, {1 + 0.1*(1-3)^2 - 0.1*(\x-3)^2 + 0.2*\x}) -- cycle;

  \foreach \y in {1.6, 2.3, 3.0, 3.7, 4.4} {
      \draw[cleangreen!40, thick] plot[domain=1:5, samples=15] (\x, \y, {1 + 0.1*(\x-3)^2 - 0.1*(\y-3)^2 + 0.2*\y});
  }
  \foreach \X in {1.6, 2.3, 3.0, 3.7, 4.4} {
      \draw[cleangreen!40, thick] plot[domain=1:5, samples=15] (\X, \x, {1 + 0.1*(\X-3)^2 - 0.1*(\x-3)^2 + 0.2*\x});
  }
  \node[cleangreen, font=\small\bfseries] at (1.0, 4.8, 0.6) {Manifold $\mathcal{M}$};

  \coordinate (X) at (3.7, 2.3, 1.46);
  \coordinate (Xtilde) at (3.7, 2.3, 2.8);
  \draw[-{Stealth[length=5pt]}, repairblue, thick, dashed] (Xtilde) -- (X)
    node[midway, left=2pt] {$f_\theta(\tilde{\bm{x}}) \approx \bm{x}$};
  \fill[cleangreen] (X) circle (2.5pt)
    node[below=3pt, fill=white, inner sep=1pt, rounded corners=1pt] {$\bm{x}$};
  \fill[corruptred] (Xtilde) circle (2.5pt)
    node[above=3pt] {$\tilde{\bm{x}} = \bm{x}+\bm{\epsilon}$};

  \coordinate (Xanom) at (1.6, 3.7, 4.8);
  \coordinate (Xproj) at (1.6, 3.7, 3.8);
  \coordinate (Xsurf) at (1.6, 3.7, 1.887);
  \draw[thick, gray, densely dotted] (Xproj) -- (Xsurf);
  \fill[cleangreen!60, draw=cleangreen, thick] (Xsurf) circle (1.5pt);
  \draw[-{Stealth[length=5pt]}, scoreorange!70, thick, dashed] (Xanom) -- (Xproj);
  \fill[scoreorange] (Xanom) circle (2.5pt)
    node[above=3pt] {$\bm{x}_\text{anom}$};
  \fill[scoreorange!20, draw=scoreorange, thick] (Xproj) circle (2pt)
    node[anchor=east, text=scoreorange, fill=white, inner sep=1pt, rounded corners=1pt, xshift=-4pt] {$f_\theta(\bm{x}_\text{anom})$};

  \draw[decorate, decoration={brace,amplitude=4pt,mirror}, thick, gray]
    ($(Xproj) + (0.35cm, 0)$) -- ($(Xanom) + (0.35cm, 0)$)
    node[midway, right=5pt, font=\small, text=gray] {score $\uparrow$};
\end{tikzpicture}%
}
% OLD: \caption{The geometric repair principle. \emph{Training} (blue dashed arrow): $f_\theta$ maps corrupted inputs $\tilde{\bm{x}}$ to clean points $\bm{x}$ on $\mathcal{M}$. \emph{Inference} (orange): anomalous $\bm{x}_\text{anom}$ far from $\mathcal{M}$ is projected toward the manifold but cannot be faithfully repaired, yielding elevated structural discrepancy.}
\caption{Repair training and inference. \emph{Training} (blue dashed arrow): the network reconstructs a clean normal window $\bm{x}$ from its corrupted version $\tilde{\bm{x}}$. \emph{Inference} (orange): the discrepancy between a test window $\bm{x}_\text{anom}$ and the network output $f_\theta(\bm{x}_\text{anom})$ contributes to the anomaly score. The surface depicts the geometric interpretation, not a guaranteed projection.}
\label{fig:manifold}
\end{figure}

This paper performs that test with \jure{} (Just Repair), a detector reduced to the essentials of the denoising principle: a single depthwise-separable convolutional residual block trained so that $f_\theta(\tilde{\bm{x}}) \approx \bm{x}$, where $\tilde{\bm{x}}$ is the input corrupted by Gaussian noise and channel masking, scored at inference by a fixed structural discrepancy function with no learned weights.
% OLD: \Cref{fig:manifold} illustrates the geometric principle: training pulls corrupted points back onto the manifold of normal windows, so at inference an anomalous window is projected toward the manifold but cannot be faithfully repaired, and the residual structure exposes it.
\Cref{fig:manifold} illustrates the repair procedure. During training, the network reconstructs clean normal windows from corrupted versions. At inference, the discrepancy between a test window and the network output is used as evidence of an anomaly.
If this minimal design were far behind the state of the art, architectural complexity would be vindicated.
% OLD: Instead, \jure{} places second on both TSB-AD-M and UCR while running roughly $20\times$ faster than the only method that ranks above it, and our ablations trace its performance to the corruption objective rather than to any architectural ingredient.
Instead, \jure{} places second on both TSB-AD-M~\cite{liu_elephant_2024} and UCR~\cite{UCRArchive,9537291}. On TSB-AD-M, it runs roughly $20\times$ faster than AxonAD~\cite{ozer_axonad_2026}, one of the top-ranked methods on that benchmark. On UCR, MatrixProfile~\cite{yeh_matrix_2016} ranks first and is also faster than \jure{}. Across the evaluated ablations, removing Gaussian corruption produces a larger AUC-PR reduction than the observed spread among the architecture variants.

%\noindent\textbf{Contributions.}
% OLD: \noindent Our contributions are:
% OLD: \begin{enumerate}
% OLD: \item \textbf{A minimal repair formulation.} \jure{} learns to map corrupted windows back to their clean form rather than to reproduce clean inputs, which discourages identity mapping and yields competitive detection with a single depthwise-separable residual block. We identify depthwise-separable convolution---separating temporal filtering from cross-channel interaction---as an effective inductive bias for this task, outperforming standard convolution and linear temporal mixing under matched conditions.
% OLD: \item \textbf{A fixed structural score.} A non-learned discrepancy function compares repaired and observed windows in amplitude, temporal differences, local trend, and cross-channel correlation, improving aggregate detection over amplitude-only scoring without introducing learned scoring parameters.
% OLD: \item \textbf{Evidence that the objective outweighs the architecture.} Across matched full-benchmark ablations on 180 series, removing Gaussian corruption degrades detection far more than any evaluated architectural substitution or increase in depth, once capacity exceeds the estimated effective dimension of the data.
% OLD: \item \textbf{Controlled validation of the geometric interpretation.} On a synthetic manifold with known projection directions, \jure{}'s scores correlate with true off-manifold distance, and its repair direction aligns increasingly with the true projection as anomaly magnitude grows.
% OLD: \end{enumerate}
\noindent Our main contribution is a deliberately minimal instantiation of corrupted-input denoising for time-series anomaly detection: a single depthwise-separable residual block paired with a fixed, non-learned score that compares test windows and network outputs in amplitude, temporal differences, local trend, and cross-channel correlation.

\section{Related Work}\label{sec:related}

% OLD: \textbf{Reconstruction and forecasting.} Autoencoders~\cite{Hinton2006}, LSTMs~\cite{malhotra_lstm_2015}, and temporal convolutional networks~\cite{tcn_2016} score by reconstruction or prediction residual. USAD~\cite{audibert_usad_2020} adds adversarial training, OmniAnomaly~\cite{su_robust_2019} uses stochastic RNNs with normalizing flows, SISVAE~\cite{li_anomaly_2021} imposes smoothness through sequential VAE priors, and Stream-VAE~\cite{ozer_stream-vae_2025} separates slow drift from fast dynamics via dual-path encoders. \jure{} belongs to this family but differs in two respects: inputs are corrupted during training, and scoring compares structure rather than raw residuals.

% OLD: \textbf{Attention and learned discrepancy.} Anomaly Transformer~\cite{xu_anomaly_2022} scores by association discrepancy over attention, TranAD~\cite{tuli_tranad_2022} combines transformers with adversarial learning, and GDN~\cite{deng_graph_2021} learns sensor graphs via attention. The current TSB-AD leader, AxonAD, measures query predictability in a JEPA-style objective~\cite{ozer_axonad_2026}. These methods demonstrate that learned discrepancy signals can capture structure beyond pure reconstruction; our results quantify how much of that structure a minimal repair objective already recovers.

Unsupervised time-series anomaly detection commonly estimates normal behavior through forecasting, reconstruction, or density modeling. Stacked LSTM predictors use forecast error~\cite{malhotra_lstm_2015}, while Donut models seasonal KPIs with a variational autoencoder~\cite{xu_unsupervised_2018}. OmniAnomaly combines stochastic recurrent states with normalizing flows~\cite{su_robust_2019}; SISVAE regularizes sequential VAE estimates for temporal smoothness~\cite{li_anomaly_2021}; and USAD couples autoencoding with adversarial training~\cite{audibert_usad_2020}. STREAM-VAE further separates slow drift from fast dynamics through dual-path routing~\cite{ozer_stream-vae_2025}. Although their architectures and probabilistic assumptions differ, these methods derive anomaly evidence from prediction, reconstruction, or likelihood-related discrepancies.

Corrupted-to-clean training has a separate lineage. Denoising autoencoders learn representations by reconstructing clean samples from corrupted inputs~\cite{vincent2008extracting,vincent2010stacked}, and their reconstruction vector field is connected to the score of a smoothed data distribution~\cite{vincent2011connection}. Score-based generative models estimate this field directly~\cite{song2020generativemodelingestimatinggradients}; CSDI applies conditional score-based diffusion to time-series imputation~\cite{tashiro2021csdiconditionalscorebaseddiffusion}; and recent work revisits clean-target prediction as a route to smaller generative models~\cite{li_jit_2025}. \jure{} does not present corrupted-input denoising as a new principle. Its methodological contribution is to instantiate that principle in a deliberately small anomaly detector and pair it with a fixed multi-component score.

\begin{figure*}[!t]
\centering
\resizebox{0.9\textwidth}{!}{%
\begin{tikzpicture}[
  every node/.style={font=\small},
  block/.style={draw, rounded corners=3pt, minimum height=0.85cm,
                minimum width=1.7cm, fill=#1, font=\small\sffamily, align=center},
  block/.default={repairblue!10},
  arr/.style={-{Stealth[length=5pt]}, thick},
  skip/.style={-{Stealth[length=5pt]}, thick, dashed, gray}
]
  \node[block=gridbg, minimum width=1.9cm] (input) at (0,0) {Input $\bm{x}$};
  \node[font=\scriptsize, below=1pt of input] {$B{\times}W{\times}C$};
  \node[block=repairblue!15, right=0.7cm of input, minimum width=1.9cm] (projin) {Conv1D $1{\times}1$};
  \node[font=\scriptsize, below=1pt of projin] {$C \to H$};
  \node[block=repairblue!30, right=0.7cm of projin,
        minimum width=3.2cm, minimum height=1.55cm, text width=3.0cm] (block1)
    {DW-Conv $5{\times}1$\\PW-Conv $1{\times}1$\\GELU + Residual};
  \node[font=\scriptsize\bfseries, above=1pt of block1] {Block};
  \node[block=repairblue!15, right=0.7cm of block1, minimum width=1.9cm] (projout) {Conv1D $1{\times}1$};
  \node[font=\scriptsize, below=1pt of projout] {$H \to C$};
  \node[font=\scriptsize, above=1pt of projout, scoreorange] {zero-init};
  \node[block=scoreorange!20, right=0.6cm of projout, minimum width=1.3cm] (delta) {$\delta_\theta$};
  \node[draw, circle, minimum size=0.55cm, right=0.55cm of delta, thick] (sum) {$+$};
  \node[block=cleangreen!20, right=0.55cm of sum, minimum width=1.9cm] (output) {$f_\theta(\bm{x})$};
  \draw[arr] (input)--(projin);
  \draw[arr] (projin)--(block1);
  \draw[arr] (block1)--(projout);
  \draw[arr] (projout)--(delta);
  \draw[arr] (delta)--(sum);
  \draw[arr] (sum)--(output);
  \draw[skip] (input.south)--++(0,-0.75)-|(sum.south);
  \node[gray, font=\scriptsize] at ($(input.south)+(3.8,-0.9)$) {global residual};
\end{tikzpicture}%
}
\caption{\jure{} architecture. A single depthwise-separable residual block between two $1{\times}1$ projections, with a global skip from input to output. The output projection is zero-initialized so $f_\theta(\bm{x})=\bm{x}$ at initialization. $H=128$ throughout.}
\label{fig:architecture}
\end{figure*}

Other time-series detectors learn structured discrepancies rather than relying only on pointwise residual magnitude. Anomaly Transformer uses attention-based association discrepancy~\cite{xu_anomaly_2022}, GDN learns a graph of inter-sensor dependencies~\cite{deng_graph_2021}, and TranAD combines transformer self-conditioning with adversarial training~\cite{tuli_tranad_2022}. AxonAD scores mismatches between predicted and target attention queries~\cite{ozer_axonad_2026}. These approaches provide learned representations of temporal or cross-channel structure. \jure{} neither reproduces nor isolates their internal mechanisms; it instead uses a fixed score combining amplitude, temporal differences, local trend, and cross-channel correlation.

Classical detectors provide an important complementary comparison. LOF measures local density deviation~\cite{breunig_lof_2000}, Isolation Forest isolates unusual observations through random partitions~\cite{liu_isolation_2008}, and COPOD estimates multivariate tail probabilities with empirical copulas~\cite{li_copod_2020}. MatrixProfile performs all-pairs subsequence similarity search~\cite{yeh_matrix_2016} and is the strongest evaluated method on UCR. Recent benchmark studies also show that rankings depend on archive construction and evaluation measure~\cite{liu_elephant_2024,9537291,paparrizos_volume_2022}. 

Against this background, \jure{} asks how far an established denoising objective, a minimal network, and a fixed score can go under standardized evaluation. The benchmark comparisons establish relative detection and efficiency tradeoffs, while the ablations characterize the evaluated design choices rather than a universal causal ordering.

\section{Just Repair (\jure{})}\label{sec:method}

Let $\bm{x} \in \RR^{W \times C}$ be a window of $W$ timesteps over $C$ channels.
% OLD: Under normal operating conditions, physical dynamics constrain windows to a manifold $\mathcal{M} \subset \RR^{W \times C}$ of effective dimension $d \ll WC$; we verify this empirically in \cref{sec:analysis}.
The manifold hypothesis models normal windows as concentrated near a lower-dimensional set $\mathcal{M} \subset \RR^{W \times C}$; \cref{sec:analysis} tests the model's capacity empirically through width and depth ablations.
% OLD: \jure{} consists of three ingredients: a repair objective that teaches the network the geometry of $\mathcal{M}$ (\cref{sec:objective}), a minimal convolutional network (\cref{sec:architecture}), and a fixed structural score (\cref{sec:scoring}).
\jure{} consists of three ingredients: a repair objective that encourages nontrivial denoising corrections (\cref{sec:objective}), a minimal convolutional network (\cref{sec:architecture}), and a fixed structural score (\cref{sec:scoring}).

\subsection{Repair Objective}\label{sec:objective}

We define \emph{repair} as learning $f_\theta$ such that $f_\theta(\tilde{\bm{x}}) \approx \bm{x}$, where $\tilde{\bm{x}} = \text{corrupt}(\bm{x})$ applies Gaussian noise ($\sigma = 0.1$) and channel masking ($p = 0.05$). The Gaussian component carries the theoretical weight: for corruption $\tilde{\bm{x}} = \bm{x} + \sigma\bm{\epsilon}$, the Bayes-optimal denoiser satisfies $f^*(\tilde{\bm{x}}) = \EE[\bm{x}|\tilde{\bm{x}}] = \tilde{\bm{x}} + \sigma^2 \nabla \log p_\sigma(\tilde{\bm{x}})$, where $p_\sigma$ is the Gaussian-smoothed data density~\cite{vincent2011connection, song2020generativemodelingestimatinggradients}.
% OLD: For small $\sigma$, this maps corrupted points toward $\mathcal{M}$, so a well-trained repair network leaves normal windows nearly unchanged while pulling anomalous windows toward the manifold. Channel masking additionally forces the network to learn cross-channel dependencies.
For small $\sigma$, the Bayes-optimal correction points toward higher-density regions under $p_\sigma$. This motivates interpreting learned corrections as movement toward regions represented by normal data, without guaranteeing such behavior for arbitrary inputs. Channel masking additionally encourages the network to learn cross-channel dependencies.
Although a tuning-subset sweep shows AUC-PR still rising modestly at $\sigma=0.2$ and $\sigma=0.4$ (\cref{sec:analysis}), we fix $\sigma=0.1$ as the smallest value near the high-performing plateau, preserving the small-noise regime on which the score-matching interpretation rests.

This objective differs from reconstruction, which trains $f_\theta(\bm{x}) \approx \bm{x}$ on clean data and therefore admits identity mapping as a trivial solution.
% OLD: Corrupting the input ($\tilde{\bm{x}} \neq \bm{x}$) forces genuine manifold learning; \cref{sec:analysis} quantifies how much detection quality this corruption contributes.
Corrupting the input ($\tilde{\bm{x}} \neq \bm{x}$) prevents the identity function from minimizing the training loss and encourages nontrivial denoising corrections; \cref{sec:analysis} reports the contribution of corruption among the evaluated variants.

The training loss combines amplitude and first-difference reconstruction:
\begin{equation}\label{eq:loss}
  \mathcal{L} = \ell_1^\text{smooth}(f_\theta(\tilde{\bm{x}}), \bm{x}) + \lambda_\Delta\,\ell_1^\text{smooth}(\Delta f_\theta(\tilde{\bm{x}}), \Delta \bm{x}),
\end{equation}
% OLD: with $\lambda_\Delta = 0.25$ encouraging preservation of temporal structure. We use Huber loss for its gradient stability near the identity initialization and robustness to anomalous-window outliers; a full-benchmark comparison against MSE and MAE appears in \cref{sec:analysis}. Optimization uses AdamW (learning rate $10^{-3}$, weight decay $10^{-4}$), batch size 128, and 30 epochs with patience-3 early stopping.
with $\lambda_\Delta = 0.25$ encouraging preservation of temporal structure. We use Huber loss for its gradient stability near the identity initialization and robustness to anomalous-window outliers; a full-benchmark comparison against MSE and MAE appears in \cref{sec:analysis}. Optimization uses AdamW (learning rate $10^{-3}$, weight decay $10^{-4}$), batch size 128, a maximum of 30 epochs, and early stopping with patience set to 3.

\subsection{Architecture}\label{sec:architecture}

The network (\cref{fig:architecture}) is a $1{\times}1$ projection to hidden width $H$, a single depthwise-separable residual block, and a zero-initialized output projection:
% OLD: \begin{align*}
% OLD:   \text{Block}(\bm{h}) &= \bm{h} + \text{GELU}(\text{PW}(\text{DW}(\bm{h}))), \quad
% OLD:   f_\theta(\bm{x}) = \bm{x} + \delta_\theta(\bm{x}).
% OLD: \end{align*}
\begin{align*}
  \text{Block}(\bm{h})
    &= \bm{h} + \text{GELU}(\text{PW}(\text{DW}(\bm{h}))), \\
  f_\theta(\bm{x})
    &= \bm{x} + \delta_\theta(\bm{x}).
\end{align*}
With $H = 128$, the parameter count is $17{,}665$ at $C = 1$. Each design choice follows from the repair task, and each is validated on the full 180-series benchmark in \cref{sec:analysis}:

% OLD: \emph{(i) Convolution.} Repair maps local temporal neighbourhoods to denoised versions. Convolution provides exactly this translation-equivariant local operation~\cite{LeCun_CNN_1989}, with a fixed receptive field matching the locality of the corruption process.
\emph{(i) Convolution.} Repair maps local temporal neighborhoods to denoised versions. Convolution provides exactly this translation-equivariant local operation~\cite{LeCun_CNN_1989}, with a fixed receptive field matching the locality of the corruption process.

% OLD: \emph{(ii) Depthwise separability.} Depthwise-separable convolution decomposes into per-channel temporal filtering (kernel size 5) and cross-channel mixing ($1{\times}1$), reducing parameters from $\mathcal{O}(H^2 K)$ to $\mathcal{O}(HK + H^2)$. Temporal smoothness and channel correlation are naturally separable aspects of repair, and the decomposition mirrors the two corruption operators: temporal filtering addresses Gaussian noise, cross-channel mixing addresses masked channels.
\emph{(ii) Depthwise separability.} Depthwise-separable convolution decomposes into per-hidden-channel temporal filtering (kernel size 5) and cross-channel mixing ($1{\times}1$), reducing parameters from $\mathcal{O}(H^2 K)$ to $\mathcal{O}(HK + H^2)$. Temporal smoothness and channel correlation are naturally separable aspects of repair, and the decomposition mirrors the two corruption operators: temporal filtering addresses Gaussian noise, cross-channel mixing addresses masked channels.

\emph{(iii) A single block.} One residual block suffices on the evaluated benchmarks; added depth yields no measurable improvement (\cref{sec:analysis}).

\emph{(iv) GELU and zero-initialization.} GELU~\cite{hendrycks2023gaussianerrorlinearunits} avoids dead neurons in the small-correction regime, and zero-initializing the output projection makes $f_\theta(\bm{x}) = \bm{x}$ exact at the start of training, so the network learns corrections from the identity.

\subsection{Structural Scoring and Inference}\label{sec:scoring}

\begin{figure}[t]
\centering
\includegraphics[width=\columnwidth]{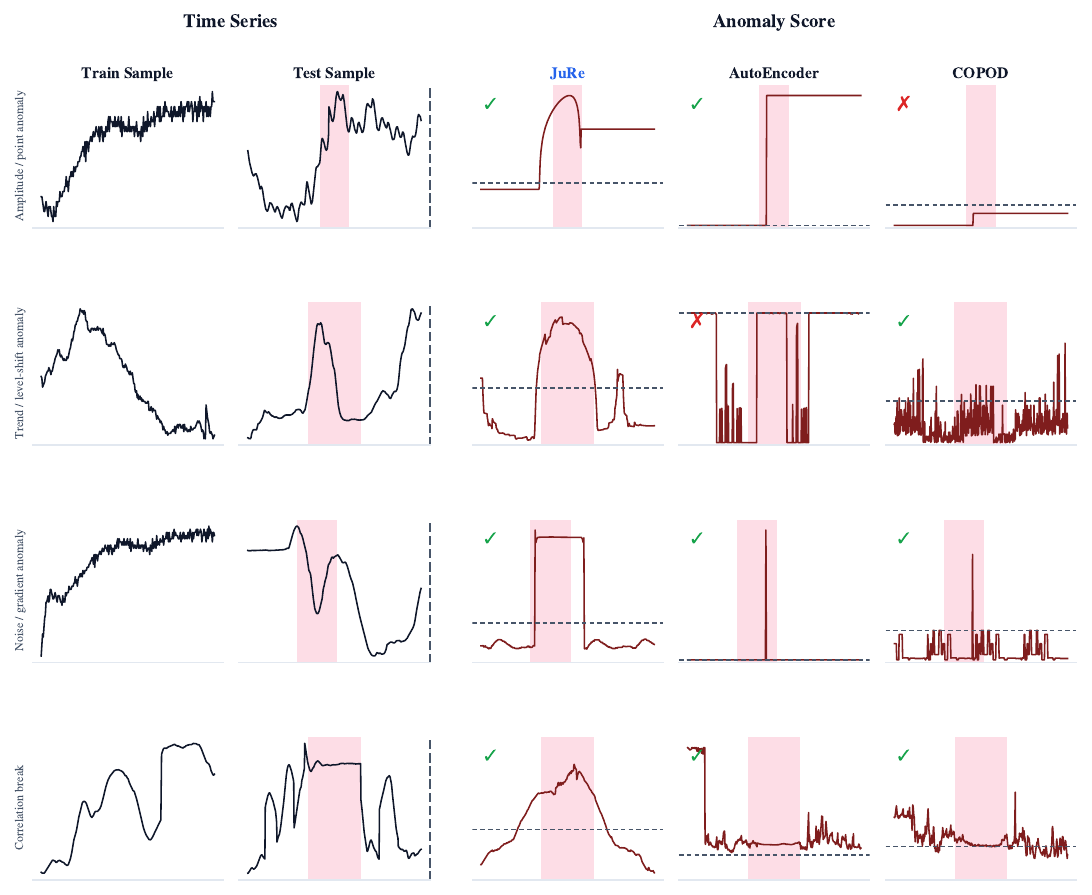}
\caption{Anomaly typology on four TSB-AD-M datasets. Each row shows a different anomaly class (amplitude spike, trend shift, gradient noise, correlation break). \emph{Left}: training snippet and test window with anomaly (pink). \emph{Right}: normalized scores for three models; dashed line marks the 95th-percentile threshold.}
\label{fig:anomaly_typology}
\end{figure}

At test time, no corruption is applied: normal windows map approximately to themselves, while anomalous windows are altered by the repair. Raw amplitude discrepancy alone, however, can miss structural anomalies with a small $\ell_1$ footprint; \cref{fig:anomaly_typology} shows four such cases spanning amplitude spikes, trend shifts, gradient noise, and correlation breaks. We therefore compare repaired and observed windows along four fixed axes:
\begin{equation}\label{eq:structure}
  s(\bm{x}) = s_\text{amp} + w_\text{diff}\,s_\text{diff} + w_\text{trend}\,s_\text{trend} + w_\text{corr}\,s_\text{corr},
\end{equation}
% OLD: where $s_\text{amp}$ is the mean absolute amplitude error, $s_\text{diff}$ the first-difference error, $s_\text{trend}$ compares moving-average trends (kernel width $\lfloor W/10 \rfloor$), and $s_\text{corr}$ is the RMS change in the upper-triangular correlation matrix (set to 0 for $C=1$; undefined entries from constant channels are omitted). The fixed weights $w_\text{diff} = 0.5$, $w_\text{trend} = 0.5$, $w_\text{corr} = 0.25$ are set by matching typical per-component magnitudes on the tuning subset so that each term contributes comparably; no scoring parameter is learned.
where $s_\text{amp}$ is the mean absolute amplitude error, $s_\text{diff}$ the first-difference error, $s_\text{trend}$ compares moving-average trends (kernel width $\lfloor W/10 \rfloor$), and $s_\text{corr}$ is the RMS change in the upper-triangular correlation matrix (set to 0 for $C=1$; undefined entries from constant channels are omitted). We use the fixed weights $w_\text{diff} = 0.5$, $w_\text{trend} = 0.5$, and $w_\text{corr} = 0.25$ throughout; no scoring parameter is learned or fitted per series. \Cref{fig:hyper_sweep} reports post-hoc sensitivity to these weights on the tuning subset.

Inference proceeds in four steps: (1) extract overlapping windows with stride 1; (2) compute $s(\bm{x})$; (3) robustly standardize using the training-set median and IQR (with $\varepsilon=10^{-8}$ added to the IQR to handle degenerate constant-score series); (4) map scores back to the time axis by averaging overlaps. \Cref{fig:pipeline} summarizes both workflows.

\begin{figure}[!b]
\centering
\resizebox{\columnwidth}{!}{%
\begin{tikzpicture}[every node/.style={font=\scriptsize}]
  \node[font=\scriptsize\bfseries, anchor=west] at (-0.5,0.65) {Training:};
  \foreach \x/\col/\lbl in {0/gridbg/{$\bm{x}$}, 2.1/corruptred!12/{$\tilde{\bm{x}}$},
                              4.2/repairblue!18/{$f_\theta(\tilde{\bm{x}})$}, 6.2/scoreorange!12/{$\mathcal{L}$}}{
    \node[draw, rounded corners=2pt, fill=\col,
          minimum width=1.55cm, minimum height=0.65cm] at (\x,0) {\lbl};
  }
  \draw[-{Stealth[length=4pt]},thick](0.78,0)--(1.32,0)
    node[midway,above,font=\tiny]{$+\epsilon$, mask};
  \draw[-{Stealth[length=4pt]},thick](2.88,0)--(3.42,0)
    node[midway,above,font=\tiny]{$f_\theta$};
  \draw[-{Stealth[length=4pt]},thick](4.98,0)--(5.42,0);
  \draw[-{Stealth[length=4pt]},dashed,gray,thick](0,-0.33)--++(0,-0.38)-|(6.2,-0.33);
  \node[gray,font=\tiny] at (3.1,-0.62) {compare to clean target};
  \node[font=\scriptsize\bfseries, anchor=west] at (-0.5,-2.1) {Inference:};
  \foreach \x/\col/\lbl in {0/gridbg/{$\bm{x}_\text{test}$},
                              2.1/repairblue!18/{$f_\theta(\bm{x})$},
                              4.5/scoreorange!12/{$s(\bm{x})$}}{
    \node[draw, rounded corners=2pt, fill=\col,
          minimum width=1.55cm, minimum height=0.65cm] at (\x,-2.75) {\lbl};
  }
  \draw[-{Stealth[length=4pt]},thick](0.78,-2.75)--(1.32,-2.75)
    node[midway,above,font=\tiny]{$f_\theta$};
  \draw[-{Stealth[length=4pt]},thick](2.88,-2.75)--(3.72,-2.75)
    node[midway,above,font=\tiny]{struct.};
  \draw[-{Stealth[length=4pt]},dashed,gray,thick](0,-3.08)--++(0,-0.38)-|(4.5,-3.08);
  \node[gray,font=\tiny] at (2.25,-3.37) {score by discrepancy};
\end{tikzpicture}%
}
\caption{Training repairs corrupted windows against clean targets. Inference scores uncorrupted test windows by structural discrepancy.}
\label{fig:pipeline}
\end{figure}

\section{Experimental Setup}\label{sec:setup}

We evaluate on the TSB-AD multivariate benchmark (180 series drawn from 17 datasets)~\cite{liu_elephant_2024} and the UCR univariate anomaly archive (250 series)~\cite{UCRArchive, 9537291}. Metrics are AUC-PR, AUC-ROC, VUS-PR, and VUS-ROC~\cite{paparrizos_volume_2022}, plus UCR-Score on the univariate archive.

\jure{} uses a single configuration throughout: $W=100$, $H=128$, $\sigma=0.1$, $p=0.05$, $\lambda_\Delta=0.25$, and score weights $(1, 0.5, 0.5, 0.25)$, with no per-dataset tuning. All hyperparameters were fixed on 20 development series drawn from TSB-AD-M but excluded from the 180-series evaluation set (the full list is provided in the repository), so all benchmark results are held-out evaluations rather than model-selection runs. The full-benchmark sweeps in \cref{sec:analysis} are post-hoc sensitivity analyses and played no role in model selection. All 25 baselines were rerun with identical data splits, preprocessing, and evaluation procedures, with method-specific hyperparameters following their published implementations. Each neural method was trained once per series with a fixed random seed; only the synthetic manifold experiment uses five seeds to report variance.

% OLD: \textbf{Statistical caveat.} Because the 180 series are nested within 17 datasets, per-series Wilcoxon tests may overstate significance for methods that perform consistently within a dataset; results should be interpreted at the series level.
Dependence among series from the same dataset may make the nominal per-series Wilcoxon $p$-values anticonservative. We therefore treat these tests as descriptive evidence across the evaluated series rather than as confirmatory evidence of generalization across independent datasets.

\section{Results}\label{sec:results}

\subsection{TSB-AD Multivariate}

% OLD: \Cref{tab:tsbad} reports the 180-series benchmark. \jure{} ranks second with AUC-PR 0.404, trailing only the attention-based AxonAD (0.437); \cref{sec:efficiency} quantifies the efficiency gap between the two. Every other baseline---including adversarial, stochastic-recurrent, and variational designs---falls below the single-block repair network.
\Cref{tab:tsbad} reports the 180-series benchmark. \jure{} ranks second with AUC-PR 0.404, trailing only the attention-based AxonAD (0.437); \cref{sec:efficiency} quantifies the efficiency gap between the two. Every other baseline, including adversarial, stochastic-recurrent, and variational designs, falls below the single-block repair network.

\begin{table}[t]
\centering
\caption{TSB-AD multivariate benchmark (180 series). Mean over all series. Best in \textbf{bold}.}\label{tab:tsbad}
\uniformtable
\begin{tabularx}{\columnwidth}{@{}Xcccc@{}}
\toprule
\textbf{Model} & \textbf{AUC-PR} & \textbf{AUC-ROC} & \textbf{VUS-PR} & \textbf{VUS-ROC} \\
\midrule
\rowcolor{repairblue!8}
\jure{} (ours) & 0.404 & 0.802 & 0.444 & 0.832 \\
\midrule
AxonAD~\cite{ozer_axonad_2026} & \textbf{0.437} & \textbf{0.825} & \textbf{0.493} & \textbf{0.859} \\
Stream-VAE~\cite{ozer_stream-vae_2025} & 0.399 & 0.808 & 0.450 & 0.838 \\
OmniAnomaly~\cite{su_robust_2019} & 0.372 & 0.744 & 0.424 & 0.777 \\
USAD~\cite{audibert_usad_2020} & 0.363 & 0.738 & 0.412 & 0.771 \\
WVAE~\cite{pereira_unsupervised_2019} & 0.354 & 0.747 & 0.413 & 0.778 \\
CNN~\cite{ren_time-series_2019} & 0.347 & 0.770 & 0.352 & 0.807 \\
VASP~\cite{von_schleinitz_vasp_2021} & 0.339 & 0.762 & 0.401 & 0.809 \\
SISVAE~\cite{li_anomaly_2021} & 0.323 & 0.759 & 0.372 & 0.786 \\
M2N2~\cite{kim2024modelmeetsnewnormals} & 0.319 & 0.740 & 0.323 & 0.779 \\
OFA~\cite{zhou2023fitsallpowergeneraltime} & 0.300 & 0.639 & 0.367 & 0.694 \\
AutoEncoder~\cite{Hinton2006} & 0.294 & 0.669 & 0.295 & 0.691 \\
VSVAE~\cite{pereira_unsupervised_2018} & 0.290 & 0.709 & 0.342 & 0.734 \\
GDN~\cite{deng_graph_2021} & 0.272 & 0.738 & 0.332 & 0.802 \\
TranAD~\cite{tuli_tranad_2022} & 0.258 & 0.675 & 0.308 & 0.742 \\
KMeansAD~\cite{yairi_kmeans_2001} & 0.252 & 0.691 & 0.296 & 0.732 \\
TFTResidual~\cite{lim_temporal_2021} & 0.250 & 0.710 & 0.308 & 0.777 \\
LSTMAD~\cite{malhotra_lstm_2015} & 0.248 & 0.597 & 0.245 & 0.626 \\
PCA~\cite{MACKIEWICZ1993303} & 0.242 & 0.676 & 0.277 & 0.712 \\
Donut~\cite{xu_unsupervised_2018} & 0.213 & 0.627 & 0.262 & 0.693 \\
IForest~\cite{liu_isolation_2008} & 0.210 & 0.704 & 0.253 & 0.750 \\
COPOD~\cite{li_copod_2020} & 0.205 & 0.652 & 0.203 & 0.686 \\
TimesNet~\cite{wu_timesnet_2023} & 0.201 & 0.618 & 0.271 & 0.686 \\
FITS~\cite{xu_fits_2024} & 0.197 & 0.611 & 0.267 & 0.686 \\
LOF~\cite{breunig_lof_2000} & 0.096 & 0.534 & 0.138 & 0.597 \\
AnomalyTr.~\cite{xu_anomaly_2022} & 0.068 & 0.506 & 0.115 & 0.538 \\
\bottomrule
\end{tabularx}
\end{table}

\Cref{tab:signif} reports pairwise Wilcoxon signed-rank tests on per-series AUC-PR against all 25 baselines. Under Holm--Bonferroni correction at family-wise $\alpha=0.05$, \jure{} is significantly superior to 20 baselines. Four comparisons do not survive correction despite \jure{}'s higher mean: Stream-VAE ($p=0.341$), OmniAnomaly ($p=0.118$), WVAE ($p=0.084$), and USAD ($p=0.018$, above the rank-22 Holm threshold of $0.05/4=0.0125$). AxonAD significantly outperforms \jure{} ($p=0.002$). All reported $p$-values are unadjusted; significance claims use the Holm thresholds throughout.

\begin{table}[t]
\centering
\caption{Pairwise Wilcoxon signed-rank tests vs.\ \jure{} on TSB-AD AUC-PR (180 series). $^\dagger$Baseline significantly outperforms \jure{} after Holm correction. \textbf{Bold}: \jure{} significantly outperforms the baseline after Holm--Bonferroni correction at family-wise $\alpha=0.05$. Reported $p$-values are unadjusted.}\label{tab:signif}
\uniformtable
\begin{tabularx}{\columnwidth}{@{}Xrrrr@{}}
\toprule
\textbf{Baseline} & \textbf{Wins} & \textbf{Losses} & \textbf{Win\%} & \textbf{$p$-value} \\
\midrule
AnomalyTr.~\cite{xu_anomaly_2022}    & 164 & 16 & 91.1 & $\mathbf{<10^{-28}}$ \\
FITS~\cite{xu_fits_2024}              & 169 & 11 & 93.9 & $\mathbf{<10^{-26}}$ \\
TimesNet~\cite{wu_timesnet_2023}      & 158 & 22 & 87.8 & $\mathbf{<10^{-23}}$ \\
LOF~\cite{breunig_lof_2000}           & 148 & 32 & 82.2 & $\mathbf{<10^{-21}}$ \\
Donut~\cite{xu_unsupervised_2018}     & 155 & 25 & 86.1 & $\mathbf{<10^{-20}}$ \\
LSTMAD~\cite{malhotra_lstm_2015}      & 149 & 31 & 82.8 & $\mathbf{<10^{-19}}$ \\
TFTResidual~\cite{lim_temporal_2021}  & 140 & 40 & 77.8 & $\mathbf{<10^{-17}}$ \\
TranAD~\cite{tuli_tranad_2022}        & 140 & 40 & 77.8 & $\mathbf{<10^{-16}}$ \\
GDN~\cite{deng_graph_2021}            & 139 & 41 & 77.2 & $\mathbf{<10^{-15}}$ \\
IForest~\cite{liu_isolation_2008}     & 138 & 42 & 76.7 & $\mathbf{<10^{-14}}$ \\
COPOD~\cite{li_copod_2020}            & 140 & 40 & 77.8 & $\mathbf{<10^{-12}}$ \\
PCA~\cite{MACKIEWICZ1993303}          & 136 & 44 & 75.6 & $\mathbf{<10^{-12}}$ \\
OFA~\cite{zhou2023fitsallpowergeneraltime} & 133 & 47 & 73.9 & $\mathbf{<10^{-9}}$ \\
M2N2~\cite{kim2024modelmeetsnewnormals} & 132 & 48 & 73.3 & $\mathbf{<10^{-8}}$ \\
KMeansAD~\cite{yairi_kmeans_2001}     & 119 & 61 & 66.1 & $\mathbf{<10^{-8}}$ \\
AutoEncoder~\cite{Hinton2006} & 129 & 51 & 71.7 & $\mathbf{<10^{-7}}$ \\
CNN~\cite{ren_time-series_2019}       & 118 & 62 & 65.6 & $\mathbf{1.5\times10^{-6}}$ \\
SISVAE~\cite{li_anomaly_2021}         & 112 & 68 & 62.2 & $\mathbf{4.2\times10^{-5}}$ \\
VASP~\cite{von_schleinitz_vasp_2021}  &  99 & 81 & 55.0 & $\mathbf{6.9\times10^{-4}}$ \\
VSVAE~\cite{pereira_unsupervised_2018}& 102 & 78 & 56.7 & $\mathbf{6.8\times10^{-4}}$ \\
USAD~\cite{audibert_usad_2020}        & 103 & 77 & 57.2 & $0.018$ \\
AxonAD$^\dagger$~\cite{ozer_axonad_2026} & 67 & 113 & 37.2 & $0.002$ \\
WVAE~\cite{pereira_unsupervised_2019} &  95 & 85 & 52.8 & $0.084$ \\
OmniAnomaly~\cite{su_robust_2019}     & 101 & 79 & 56.1 & $0.118$ \\
Stream-VAE~\cite{ozer_stream-vae_2025} & 81 & 99 & 45.0 & $0.341$ \\
\bottomrule
\end{tabularx}
\end{table}

\subsection{UCR Univariate}

% OLD: On the univariate archive (\cref{tab:ucr}), \jure{} achieves AUC-PR 0.201, ranking second behind MatrixProfile (0.292) and first among neural methods on both AUC-PR and VUS-PR. Notably, the multivariate ranking inverts here: the attention-based AxonAD drops to 0.134, well below \jure{}, suggesting its advantage on TSB-AD stems from modelling cross-channel structure that univariate data lacks. MatrixProfile's lead is expected---it computes exact nearest-neighbour distances over all subsequences---but the evaluated implementation supports only univariate input and could not be applied to TSB-AD-M. On UCR-Score, three neural baselines rank above \jure{} (0.368); because UCR-Score rewards only the single highest-scored point, it can diverge from continuous ranking quality when a method occasionally mislocates its peak.
On the univariate archive (\cref{tab:ucr}), \jure{} achieves AUC-PR 0.201, ranking second behind MatrixProfile (0.292) and first among neural methods on both AUC-PR and VUS-PR. Notably, the multivariate ranking inverts here: the attention-based AxonAD drops to 0.134, well below \jure{}. This reversal is consistent with AxonAD benefiting from cross-channel structure that is absent in univariate data, although the benchmark comparison does not isolate that mechanism. MatrixProfile computes exact nearest-neighbor distances over all subsequences; however, the evaluated implementation supports only univariate input and could not be applied to TSB-AD-M. On UCR-Score, three neural baselines rank above \jure{} (0.368); because UCR-Score rewards only the single highest-scored point, it can diverge from continuous ranking quality when a method occasionally mislocates its peak.

\begin{table}[t]
\centering
\caption{UCR anomaly archive (250 univariate series). Best in \textbf{bold}.}\label{tab:ucr}
\uniformtable
\scriptsize
\setlength{\tabcolsep}{2.5pt}
\begin{tabularx}{\columnwidth}{@{}>{\raggedright\arraybackslash}p{2.25cm}*{5}{>{\centering\arraybackslash}X}@{}}
\toprule
\textbf{Model} & \textbf{AUC-PR} & \textbf{AUC-ROC} & \textbf{VUS-PR} & \textbf{VUS-ROC} & \textbf{UCR} \\
\midrule
\rowcolor{repairblue!8}
\jure{} (ours) & 0.201 & 0.674 & 0.199 & 0.719 & 0.368 \\
\midrule
MatrixProfile~\cite{yeh_matrix_2016} & \textbf{0.292} & \textbf{0.872} & \textbf{0.332} & \textbf{0.888} & \textbf{0.548} \\
Stream-VAE~\cite{ozer_stream-vae_2025} & 0.171 & 0.761 & 0.179 & 0.804 & 0.308 \\
KMeansAD~\cite{yairi_kmeans_2001} & 0.150 & 0.716 & 0.135 & 0.774 & 0.276 \\
AxonAD~\cite{ozer_axonad_2026} & 0.134 & 0.677 & 0.139 & 0.753 & 0.424 \\
CNN~\cite{ren_time-series_2019} & 0.100 & 0.663 & 0.082 & 0.748 & 0.428 \\
LSTMAD~\cite{malhotra_lstm_2015} & 0.088 & 0.629 & 0.058 & 0.718 & 0.392 \\
USAD~\cite{audibert_usad_2020} & 0.072 & 0.552 & 0.070 & 0.626 & 0.188 \\
IForest~\cite{liu_isolation_2008} & 0.059 & 0.595 & 0.052 & 0.651 & 0.116 \\
LOF~\cite{breunig_lof_2000} & 0.057 & 0.590 & 0.047 & 0.690 & 0.212 \\
AutoEncoder~\cite{Hinton2006} & 0.056 & 0.571 & 0.058 & 0.651 & 0.236 \\
OmniAnomaly~\cite{su_robust_2019} & 0.021 & 0.522 & 0.016 & 0.619 & 0.132 \\
TimesNet~\cite{wu_timesnet_2023} & 0.021 & 0.538 & 0.019 & 0.629 & 0.220 \\
TranAD~\cite{tuli_tranad_2022} & 0.017 & 0.508 & 0.015 & 0.614 & 0.132 \\
Donut~\cite{xu_unsupervised_2018} & 0.015 & 0.499 & 0.014 & 0.604 & 0.096 \\
AnomalyTr.~\cite{xu_anomaly_2022} & 0.012 & 0.518 & 0.018 & 0.556 & 0.168 \\
\bottomrule
\end{tabularx}
\end{table}

\subsection{Efficiency}\label{sec:efficiency}

% OLD: \jure{} has $H^2+2HC+8H+C$ parameters: 17{,}665 at $C=1$ and a median of 19{,}978 across TSB-AD-M ($C=10$). Throughput was measured on a single NVIDIA A100 GPU (batch size 256, float32, mean over all benchmark series after 50 warm-up iterations, including network inference and score computation but excluding disk I/O). \jure{} processes 9{,}870 scores/s on TSB-AD-M; AxonAD, with 358{,}916 parameters, processes 497 scores/s. The top of the leaderboard thus trades roughly $20\times$ throughput for 0.033 AUC-PR. \Cref{fig:cost_of_complexity} plots AUC-PR against parameter count and throughput for all methods---neither small nor large parameter budgets predict detection quality---and \cref{tab:throughput} lists throughput on both benchmarks.
\jure{} has $H^2+2HC+8H+C$ parameters: 17{,}665 at $C=1$ and 19{,}978 at the median TSB-AD-M channel count, $C=10$. Throughput was measured on a single NVIDIA A100 GPU (batch size 256, float32, mean over all benchmark series after 50 warm-up iterations, including network inference and score computation but excluding disk I/O). \jure{} processes 9{,}870 scores/s on TSB-AD-M; AxonAD, with 358{,}916 parameters, processes 497 scores/s. The top of the leaderboard thus trades roughly $20\times$ throughput for 0.033 AUC-PR. \Cref{fig:cost_of_complexity} plots AUC-PR against parameter count and throughput for all methods. Neither small nor large parameter budgets predict detection quality. \Cref{tab:throughput} lists throughput on both benchmarks.

\begin{figure*}[t]
\centering
\begin{tikzpicture}
\begin{groupplot}[
  group style={
    group size=2 by 1,
    horizontal sep=2.2cm,
    group name=costplots,
  },
  grid=major, grid style={gray!15},
  tick label style={font=\scriptsize},
  label style={font=\small},
  width=0.46\textwidth, height=6.2cm,
  ymin=0.0, ymax=0.50,
  clip=false,
]
\nextgroupplot[
  title={\small\bfseries Parameters vs.\ AUC-PR (TSB-AD)},
  xlabel={Trainable parameters},
  ylabel={AUC-PR},
  xmode=log, xmin=100, xmax=1.5e7,
  legend to name=costlegend,
  legend style={legend columns=2, font=\scriptsize,
                column sep=0.8em, draw=gray!40},
]
\addplot[only marks, mark=square*, mark size=2.5pt, draw=gray!70, fill=gray!40]
  coordinates {
    (4752409, 0.068)
    (2561426, 0.399)
    (2498306, 0.294)
    (814081,  0.300)
    (683962,  0.323)
    (628193,  0.250)
    (358916,  0.437)
    (283158,  0.354)
    (178454,  0.290)
    (145539,  0.339)
    (73449,   0.201)
    (20736,   0.213)
    (13717,   0.372)
    (10421,   0.248)
    (7289,    0.347)
    (7041,    0.272)
    (6109,    0.363)
    (1248,    0.197)
    (369,     0.258)
    (322,     0.319)
  };
\addlegendentry{Baselines}
\addplot[only marks, mark=*, mark size=5pt, draw=repairblue, fill=repairblue]
  coordinates {(19978, 0.404)};
\addlegendentry{\jure{} (ours)}

\node[font=\tiny, gray, anchor=south]      at (axis cs:358916,0.448)  {AxonAD};
\node[font=\tiny, gray, anchor=south east] at (axis cs:322,0.325)     {M2N2};
\node[font=\tiny, repairblue, anchor=south]at (axis cs:19978,0.418)   {\jure{}};

\nextgroupplot[
  title={\small\bfseries Throughput vs.\ AUC-PR (TSB-AD)},
  xlabel={Throughput (scores/s)},
  ylabel={AUC-PR},
  xmode=log, xmin=50, xmax=4e6,
]
\addplot[only marks, mark=square*, mark size=2.5pt, draw=gray!70, fill=gray!40]
  coordinates {
    (394,   0.068)
    (2596,  0.294)
    (8935,  0.347)
    (1846,  0.213)
    (944,   0.197)
    (3048,  0.272)
    (16788, 0.319)
    (1485,  0.248)
    (107,   0.300)
    (198,   0.372)
    (490,   0.323)
    (286,   0.250)
    (291,   0.201)
    (1463,  0.258)
    (5902,  0.363)
    (684,   0.339)
    (583,   0.354)
    (526,   0.290)
    (497,   0.437)
    (149,   0.399)
    (1020142, 0.205)
    (243752,  0.096)
    (56417,   0.252)
    (32756,   0.242)
    (29653,   0.210)
  };
\addplot[only marks, mark=*, mark size=5pt, draw=repairblue, fill=repairblue]
  coordinates {(9870, 0.404)};

\node[font=\tiny, gray, anchor=south]      at (axis cs:497,0.448)    {AxonAD};
\node[font=\tiny, gray, anchor=south]      at (axis cs:1020142,0.215){COPOD};
\node[font=\tiny, repairblue, anchor=south]at (axis cs:9870,0.418)   {\jure{}};

\end{groupplot}
\node[anchor=north] at
  ($(costplots c1r1.south)!0.5!(costplots c2r1.south)+(0,-0.65cm)$)
  {\pgfplotslegendfromname{costlegend}};
\end{tikzpicture}
% OLD: \caption{Trainable parameter count (left) and inference throughput (right) vs.\ AUC-PR on TSB-AD. Neither small nor large parameter budgets predict AUC-PR. \jure{} is roughly $20\times$ faster than AxonAD at a cost of 0.033 AUC-PR.}
\caption{Trainable parameter count (left) and inference throughput (right) vs.\ AUC-PR on TSB-AD. Labels identify representative Pareto-front methods. Neither small nor large parameter budgets predict AUC-PR. \jure{} is roughly $20\times$ faster than AxonAD at a cost of 0.033 AUC-PR.}
\label{fig:cost_of_complexity}
\end{figure*}

\begin{table}[t]
\centering
\caption{Inference throughput (scores/s) on TSB-AD-M and UCR. N/A: evaluated MatrixProfile implementation supports only univariate input.}\label{tab:throughput}
\uniformtable
\begin{tabularx}{\columnwidth}{@{}Xrr@{}}
\toprule
\textbf{Model} & \textbf{M (sc/s)} & \textbf{UCR (sc/s)} \\
\midrule
COPOD~\cite{li_copod_2020}             & 1{,}020{,}142 & 3{,}112{,}523 \\
LOF~\cite{breunig_lof_2000}           & 243{,}752 & 394{,}716 \\
KMeansAD~\cite{yairi_kmeans_2001}     & 56{,}417 & 82{,}117 \\
PCA~\cite{MACKIEWICZ1993303}          & 32{,}756 & 691{,}127 \\
IForest~\cite{liu_isolation_2008}     & 29{,}653 & 95{,}849 \\
MatrixProfile~\cite{yeh_matrix_2016}  & N/A      & 103{,}304 \\
M2N2~\cite{kim2024modelmeetsnewnormals} & 16{,}788 & 3{,}645 \\
\rowcolor{repairblue!8}
\jure{} (ours)                         & 9{,}870  & 5{,}327 \\
CNN~\cite{ren_time-series_2019}        & 8{,}935  & 4{,}281 \\
USAD~\cite{audibert_usad_2020}         & 5{,}902  & 4{,}823 \\
GDN~\cite{deng_graph_2021}             & 3{,}048  & 1{,}886 \\
AutoEncoder~\cite{Hinton2006}      & 2{,}596  & 1{,}818 \\
Donut~\cite{xu_unsupervised_2018}      & 1{,}846  & 2{,}770 \\
LSTMAD~\cite{malhotra_lstm_2015}       & 1{,}485  &   759 \\
TranAD~\cite{tuli_tranad_2022}         & 1{,}463  &   756 \\
FITS~\cite{xu_fits_2024}               &   944   & 1{,}930 \\
VASP~\cite{von_schleinitz_vasp_2021}   &   684   &   338 \\
WVAE~\cite{pereira_unsupervised_2019}  &   583   &   290 \\
VSVAE~\cite{pereira_unsupervised_2018} &   526   &   225 \\
AxonAD~\cite{ozer_axonad_2026}         &   497   &   566 \\
SISVAE~\cite{li_anomaly_2021}          &   490   &   262 \\
AnomalyTr.~\cite{xu_anomaly_2022}      &   394   &   188 \\
TimesNet~\cite{wu_timesnet_2023}       &   291   &   193 \\
TFTResidual~\cite{lim_temporal_2021}   &   286   &   403 \\
OmniAnomaly~\cite{su_robust_2019}      &   198   &   219 \\
Stream-VAE~\cite{ozer_stream-vae_2025}  &   149   &   145 \\
OFA~\cite{zhou2023fitsallpowergeneraltime} & 107 &    82 \\
\bottomrule
\end{tabularx}
\end{table}

% OLD: \section{Why Repair Works}\label{sec:analysis}
\section{Analysis of the Repair Mechanism}\label{sec:analysis}

Having established that a minimal repair network is competitive, we now dissect where its performance comes from. All analyses in this section use the full 180-series TSB-AD benchmark unless noted; hyperparameter sweeps on the 20-series tuning subset (\cref{fig:hyper_sweep}) are shown for context only.

% OLD: \subsection{The Objective Dominates}
\subsection{Effect of the Training Objective}

% OLD: \Cref{tab:ablation} reports component ablations. The largest drop among the evaluated training and scoring choices comes from removing Gaussian noise corruption while retaining channel masking: $-0.045$ AUC-PR.
\Cref{tab:ablation} reports component ablations. The largest drop among the evaluated training and scoring choices comes from removing Gaussian noise corruption while retaining channel masking: $-0.046$ AUC-PR.
% OLD: Gaussian denoising is thus the dominant manifold-learning signal, beyond what masking alone provides---direct evidence that repair, not reconstruction, drives detection.
This result shows that Gaussian corruption contributes more than channel masking alone under the evaluated conditions. Because the ablation does not include a matched clean-input condition without either corruption type, it does not isolate repair from reconstruction.
% OLD: Removing channel masking costs 0.024, while the auxiliary difference loss contributes only marginally on aggregate: the corruption design, not the loss decomposition, does the work.
Removing channel masking costs 0.024, while removing the auxiliary difference loss changes aggregate AUC-PR by 0.001.

Structural scoring contributes a second, smaller layer: amplitude-only scoring drops to 0.388, and removing the correlation term alone costs 0.023, confirming that cross-channel structure carries signal a raw residual misses (cf.~\cref{fig:anomaly_typology}). The sensitivity sweep (\cref{fig:hyper_sweep}, right) shows the difference and trend weights are robust across the evaluated range, while excessive correlation weighting ($w_\text{corr}\geq1.0$) degrades performance.

\begin{table}[t]
\centering
\caption{Component ablation on TSB-AD (180 series). \textbf{Bold} marks the reference configuration, not the column maximum.}\label{tab:ablation}
\uniformtable
\begin{tabularx}{\columnwidth}{@{}Xcc@{}}
\toprule
\textbf{Variant} & \textbf{AUC-PR} & \textbf{AUC-ROC} \\
\midrule
\jure{} (full) & \textbf{0.404} & \textbf{0.802} \\
\midrule
\textit{Scoring:} & & \\
\quad Amplitude only ($s_\text{amp}$) & 0.388 & 0.783 \\
\quad No correlation ($s_\text{corr}=0$) & 0.381 & 0.780 \\
\quad No gradient ($s_\text{diff}=0$) & 0.399 & 0.797 \\
\midrule
\textit{Training:} & & \\
\quad No Gaussian noise ($\sigma=0$, masking retained) & 0.358 & 0.757 \\
\quad No masking ($p=0$) & 0.380 & 0.796 \\
\quad No diff loss ($\lambda_\Delta=0$) & 0.403 & 0.794 \\
\midrule
\textit{Architecture:} & & \\
\quad Two blocks & 0.401 & 0.800 \\
\quad No zero-init & 0.390 & 0.798 \\
\quad $H=64$ & 0.405 & 0.801 \\
\quad $H=8$ & 0.357 & 0.770 \\
\bottomrule
\end{tabularx}
\end{table}

\begin{figure*}[t]
\centering
\resizebox{\textwidth}{!}{%
\begin{tikzpicture}
\begin{groupplot}[
    group style={
        group size=3 by 1,
        horizontal sep=1.8cm,
        group name=sweepplots,
    },
    height=4.5cm, width=0.38\textwidth,
    grid=major, grid style={gray!15},
    tick label style={font=\scriptsize},
    label style={font=\small},
    ylabel={AUC-PR},
    ymin=0.3, ymax=0.5,
    enlarge x limits=0.05
]

\nextgroupplot[
    title={\small\bfseries Noise Scale ($\sigma$)},
    xlabel={$\sigma$},
]
\addplot[mark=*, mark size=1.5pt, draw=repairblue, thick] coordinates {
    (0.0, 0.331) (0.025, 0.400) (0.05, 0.430)
    (0.1, 0.458) (0.2, 0.477) (0.4, 0.478)
};
\node[coordinate, pin={[pin distance=0.4cm, pin edge={black, thin}]270:{\tiny Baseline (0.1)}}] at (axis cs:0.1, 0.458) {};

\nextgroupplot[
    title={\small\bfseries Diff.\ Loss Weight ($\lambda_\Delta$)},
    xlabel={$\lambda_\Delta$},
]
\addplot[mark=square*, mark size=1.5pt, draw=cleangreen, thick] coordinates {
    (0.0, 0.438) (0.05, 0.448) (0.1, 0.437)
    (0.25, 0.458) (0.5, 0.440) (1.0, 0.452)
};
\node[coordinate, pin={[pin distance=0.4cm, pin edge={black, thin}]0:{\tiny Baseline (0.25)}}] at (axis cs:0.25, 0.458) {};

\nextgroupplot[
    title={\small\bfseries Score Component Weights},
    xlabel={Weight Value},
    legend style={at={(0.95,0.05)}, anchor=south east, font=\tiny,
                  draw=gray!40, fill=white, fill opacity=0.8, text opacity=1},
]
\addplot[mark=triangle*, mark size=1.8pt, draw=scoreorange, thick] coordinates {
    (0.0, 0.457) (0.25, 0.460) (0.5, 0.458) (1.0, 0.458) (2.0, 0.446)
};
\addlegendentry{$w_\text{diff}$}

\addplot[mark=diamond*, mark size=1.8pt, draw=corruptred, thick] coordinates {
    (0.0, 0.444) (0.25, 0.456) (0.5, 0.458) (1.0, 0.463) (2.0, 0.465)
};
\addlegendentry{$w_\text{trend}$}

\addplot[mark=*, mark size=1.5pt, draw=purple, thick] coordinates {
    (0.0, 0.468) (0.125, 0.467) (0.25, 0.458) (0.5, 0.467) (1.0, 0.412)
};
\addlegendentry{$w_\text{corr}$}

\end{groupplot}
\end{tikzpicture}%
}
\caption{Hyperparameter sensitivity on the 20-series tuning subset. Removing noise ($\sigma=0$, left) causes a severe drop, while the diff-loss weight (center) and the difference and trend score weights (right) are robust across the evaluated range; excessive correlation weighting ($w_\text{corr}\geq1.0$) degrades performance.}
\label{fig:hyper_sweep}
\end{figure*}

% OLD: \subsection{Capacity: How Much Network Is Needed}
\subsection{Model Capacity}

% OLD: If repair approximates projection onto $\mathcal{M}$, the hidden width $H$ needs only to exceed the effective dimension of the data.
The manifold interpretation motivates testing whether performance saturates once the network has moderate capacity. We examine this question directly through the reported width and depth ablations.
% OLD: PCA on training windows of 20 representative TSB-AD-M series covering all 17 constituent datasets (where a dataset contains multiple predefined subsets, the series nearest the subset's median channel count was selected) yields a median $d_{95}=22$ components for 95\% variance (mean 76), with $d_{95} < 128$ on 16 of 20 series---far below the ambient dimension $WC = 100C$. The exceptions are high-dimensional MSL/SMAP series.

% OLD: The ablation matches this picture precisely. At $H=8$, far below the estimated effective dimension, performance collapses ($0.357$). Between $H=64$ and $H=128$ it saturates: mean AUC-PR 0.405 vs.\ 0.404, with a paired Wilcoxon test confirming no significant difference ($W=7149$, $p=0.155$; $H=128$ wins 104 of 180 series vs.\ 76). A second block likewise adds nothing (0.401). $H=128$ is thus a pre-specified default inside a saturation regime, not a tuned optimum.
Performance is lower at $H=8$ (AUC-PR 0.357) and similar at $H=64$ and $H=128$ (0.405 vs.\ 0.404); a paired Wilcoxon test finds no significant difference between the latter two settings ($W=7149$, $p=0.155$; $H=128$ wins 104 of 180 series vs.\ 76). A second block yields AUC-PR 0.401. These observations indicate saturation across the evaluated moderate-width settings, but they do not establish a general capacity threshold. We therefore retain $H=128$ as the pre-specified default rather than treating it as a theoretically derived value.

% OLD: \subsection{Architecture Matters Less---but Its Bias Still Shows}
\subsection{Architecture and Inductive Bias}

% OLD: \Cref{tab:arch_match} compares depthwise-separable convolution against standard convolution, linear temporal mixing, and a ReLU variant at matched parameter counts. DW-sep leads standard convolution by $+0.016$ AUC-PR and the linear MLP by $+0.010$; replacing GELU with ReLU trades $-0.006$ AUC-PR for $+0.003$ AUC-ROC. Two observations follow.
\Cref{tab:arch_match} compares the evaluated depthwise-separable convolution, standard convolution, linear temporal mixing, and ReLU variants. DW-sep leads standard convolution by $+0.016$ AUC-PR and the linear MLP by $+0.010$; replacing GELU with ReLU trades $-0.006$ AUC-PR for $+0.003$ AUC-ROC. Two observations follow.
% OLD: First, the total spread across matched architectures is at most 0.017---roughly a third of the cost of removing Gaussian corruption---quantifying the claim that the objective outweighs the architecture within a non-degenerate capacity range.
First, the total spread across the evaluated architecture variants is at most 0.017, roughly a third of the 0.046 reduction observed when Gaussian corruption is removed. Because exact parameter comparability is not established here, this is a descriptive comparison of the tested variants rather than a controlled estimate of architecture's effect.
Second, the DW-sep advantage was invisible on the 20-series tuning subset and resolved only at 180 series, underscoring the value of full-benchmark ablation. The same holds for the loss comparison (\cref{tab:loss_ablation}): Huber edges out MSE (0.404 vs.\ 0.400) with MAE well behind (0.371), an ordering that differed on smaller subsets.

\begin{table}[t]
\centering
% OLD: \caption{Matched architecture comparison on the full TSB-AD benchmark (180 series).}\label{tab:arch_match}
\caption{Architecture-variant comparison on the full TSB-AD benchmark (180 series).}\label{tab:arch_match}
\uniformtable
\begin{tabularx}{\columnwidth}{@{}Xcc@{}}
\toprule
\textbf{Variant} & \textbf{AUC-PR} & \textbf{AUC-ROC} \\
\midrule
\textbf{\jure{}} (DW-sep, GELU, k=5) & \textbf{0.404} & 0.802 \\
% OLD: Standard Conv1D (matched params) & 0.387 & 0.787 \\
Standard Conv1D & 0.387 & 0.787 \\
% OLD: Linear temporal MLP (matched params) & 0.394 & 0.785 \\
Linear temporal MLP & 0.394 & 0.785 \\
DW-sep + ReLU & 0.398 & \textbf{0.805} \\
\bottomrule
\end{tabularx}
\end{table}

\begin{table}[t]
\centering
\caption{Loss function comparison on the full TSB-AD benchmark (180 series).}\label{tab:loss_ablation}
\uniformtable
\begin{tabularx}{\columnwidth}{@{}Xcc@{}}
\toprule
\textbf{Loss} & \textbf{AUC-PR} & \textbf{AUC-ROC} \\
\midrule
MSE & 0.400 & \textbf{0.805} \\
MAE & 0.371 & 0.787 \\
\textbf{Huber} (default) & \textbf{0.404} & 0.802 \\
\bottomrule
\end{tabularx}
\end{table}

% OLD: \subsection{Window Size}
\subsection{Window-Length Sensitivity}

% OLD: A sweep over $W \in \{25, 50, 75, 100, 150, 200\}$ (\cref{tab:window_size}) shows $W=100$ achieves the highest AUC-PR among the evaluated values, with graceful degradation on either side. AUC-ROC keeps rising at larger windows, which capture longer-range structure, while AUC-PR favours tighter localization; the default $W=100$ balances both without per-dataset tuning.
A sweep over $W \in \{25, 50, 75, 100, 150, 200\}$ (\cref{tab:window_size}) shows $W=100$ achieves the highest AUC-PR among the evaluated values, with graceful degradation on either side. AUC-ROC keeps rising at larger windows, which capture longer-range structure, while AUC-PR favors tighter localization; the default $W=100$ balances both without per-dataset tuning.

\begin{table}[t]
\centering
\caption{Window size sensitivity on the full TSB-AD benchmark (180 series).}\label{tab:window_size}
\uniformtable
\begin{tabularx}{\columnwidth}{@{}>{\centering\arraybackslash}X>{\centering\arraybackslash}X>{\centering\arraybackslash}X@{}}
\toprule
\textbf{$W$} & \textbf{AUC-PR} & \textbf{AUC-ROC} \\
\midrule
25  & 0.352 & 0.739 \\
50  & 0.373 & 0.767 \\
75  & 0.386 & 0.792 \\
\rowcolor{repairblue!8}
100 & \textbf{0.404} & 0.802 \\
150 & 0.391 & \textbf{0.817} \\
200 & 0.375 & \textbf{0.817} \\
\bottomrule
\end{tabularx}
\end{table}

% OLD: \subsection{Does \jure{} Actually Project onto the Manifold?}\label{sec:manifold_val}
\subsection{Synthetic Validation of the Manifold Interpretation}\label{sec:manifold_val}

Benchmark scores cannot distinguish whether the network truly approximates manifold projection or merely produces useful residuals. We therefore construct a controlled experiment with known ground truth: a $d=4$-dimensional subspace embedded in $\RR^{W \times C}$ ($W=100$, $C=10$) defines $\mathcal{M}$; 5{,}000 training windows are sampled from $\mathcal{M}$ with Gaussian noise ($\sigma_{\text{data}}=0.05$); test anomalies are planted at seven orthogonal distances $\delta \in \{0, 0.1, 0.3, 0.5, 1.0, 2.0, 3.0\}$.

Two quantities are measured over 5 seeds. First, the correlation between \jure{}'s anomaly score and the true orthogonal distance: Pearson $r = 0.725 \pm 0.004$ and Spearman $\rho = 0.611 \pm 0.009$ ($p < 10^{-219}$); the gap between the two indicates mild nonlinearity in the score--distance relationship, consistent with the structural score combining multiple signal components. Second, the cosine similarity between the repair direction $f_\theta(\bm{x}) - \bm{x}$ and the true projection direction $\Pi_{\mathcal{M}}(\bm{x}) - \bm{x}$ rises monotonically with anomaly magnitude: $0.06$ at $\delta=0.1$, $0.18$ at $\delta=0.3$, $0.50$ at $\delta=1.0$, and $0.85$ at $\delta=3.0$.
% OLD: Near the manifold, alignment is weak---the score-matching approximation is least precise there and noise dominates the displacement---but for the larger anomalies that matter most in detection, the learned repair closely tracks the true projection. This constitutes direct, if partial, support for the geometric interpretation of \cref{fig:manifold}.
Near the manifold, the true projection vector is small, making directional cosine measurements noise-sensitive and potentially ill-conditioned. At the larger tested displacements, the higher cosine similarity provides partial support for the geometric interpretation in this controlled linear setting; it does not establish projection behavior for general manifolds or real anomalies.

\section{Discussion and Limitations}\label{sec:discussion}

%\textbf{What the results say.}
% OLD: Taken together, the ablations and benchmarks support a single conclusion: within a non-degenerate capacity range, the denoising objective---not architectural complexity---drives detection quality. The corruption ablation isolates the objective's contribution, the matched architecture comparison bounds the architecture's, and the capacity analysis explains why a single narrow block suffices.
The experiments compare the evaluated corruption and architecture variants but do not establish a general causal ordering between objective and architecture.
% OLD: The two benchmarks then delimit where complexity does pay: on TSB-AD, with rich cross-channel interactions, AxonAD's attention objective opens a gap that local repair cannot close---at a steep inference cost (\cref{sec:efficiency}). On UCR, where channels are absent and temporal structure dominates, the ranking reverses and \jure{} leads all neural methods. Practitioners can read this as a deployment guideline: repair where latency matters, attention where the last decimal of multivariate accuracy does.
The two benchmarks delimit the observed tradeoffs. On TSB-AD, \jure{} provides roughly $20\times$ the throughput of AxonAD with 0.033 lower AUC-PR. On UCR, \jure{} leads all neural methods but trails MatrixProfile in both AUC-PR and throughput.

% OLD: \textbf{Limitations.} Each neural method is trained with a single fixed random seed on the real benchmarks, so multi-seed variance is characterized only in the synthetic experiment. The corruption parameters, scoring weights, and window size are fixed globally rather than tuned per dataset; the window sweep shows graceful degradation, but per-dataset tuning could recover further gains. The corruption ablation does not include a matched clean-input reconstruction condition without Gaussian corruption or channel masking.
%\textbf{Limitations.} 
Each neural method is trained with a single fixed random seed on the real benchmarks, so multi-seed variance is characterized only in the synthetic experiment. The corruption parameters, scoring weights, and window size are fixed globally rather than tuned per dataset; the window sweep shows graceful degradation, but per-dataset tuning could recover further gains. The corruption ablation does not include a matched clean-input reconstruction condition without Gaussian corruption or channel masking, and exact parameter comparability among the architecture variants is not established.
% OLD: On the few high-dimensional MSL/SMAP series where $d_{95} > 128$, the fixed hidden width may underfit.
The synthetic manifold is a linear subspace, and repair-direction alignment is weak at small displacements ($\delta \leq 0.3$); nonlinear manifolds and small-displacement robustness remain untested. Per-series significance tests may overstate confidence due to dataset nesting (\cref{sec:setup}). Finally, \jure{} operates in a batch setting; online adaptation is a natural extension.

\section{Conclusion}\label{sec:conclusion}

% OLD: We asked whether the denoising principle, implemented minimally, can match the detection quality of far more complex systems.
% OLD: On two standardized benchmarks the answer is largely yes: \jure{}, a single depthwise-separable residual block with a fixed scoring function, ranks second on both TSB-AD-M and UCR, leads all neural baselines on the univariate archive, and runs an order of magnitude faster than the only method that outperforms it.
We asked whether a minimally implemented denoising principle can remain competitive with substantially more complex systems. Across two standardized benchmarks, the results show that it can: \jure{}, a single depthwise-separable residual block with a fixed scoring function, ranks second on both TSB-AD-M and UCR and leads all neural baselines on the univariate archive. On TSB-AD-M, \jure{} runs roughly $20\times$ faster than AxonAD at a cost of 0.033 AUC-PR; on UCR, MatrixProfile remains both more accurate and faster.
% OLD: Full-benchmark ablations attribute this performance to the corruption objective rather than the architecture, and a controlled manifold experiment corroborates the geometric interpretation at the anomaly magnitudes that matter for detection.
The evaluated corruption and architecture variants do not establish a general causal ordering between objective and architecture.
% OLD: A controlled manifold experiment provides evidence for the geometric interpretation at larger anomaly magnitudes.
A controlled linear-manifold experiment provides partial evidence for the geometric interpretation at larger tested displacements.
These conclusions hold for the evaluated benchmarks; online detection, streaming, and nonlinear manifolds remain open directions.

\bibliographystyle{IEEEtran}
\bibliography{references}

\end{document}